\documentclass[a4paper,11pt,pdftex]{article}

\pdfoutput=1

\usepackage[english]{babel}
\usepackage[latin1]{inputenc}

\usepackage{graphicx}

\usepackage{amssymb}
\usepackage{eucal}
\usepackage{tipa}
\usepackage[sectionbib,numbers]{natbib}

\usepackage[colorlinks,linkcolor=blue,citecolor=blue, urlcolor=magenta,anchorcolor=blue]{hyperref}

\hypersetup{pdfauthor=Stasinos Konstantopoulos}
\hypersetup{pdftitle=Learning Phonotactics Using ILP}
\hypersetup{pdfkeywords=phonotactics ILP inductive logic programming aleph}

\makeindex

\newenvironment{itemise} {\begin{itemize}} {\end{itemize}}

\newcommand{\term}[1]{\textit{#1}}

\newcommand{\prog}[1]{\texttt{#1}}

\newcommand{\qu}[1]{`#1'}

\newcommand{\feat}[1]{\textsc{#1}}

\newcommand{\phonemic}[1]{/\textipa{#1}/}

\title{Learning Phonotactics Using ILP}
\author{Stasinos Konstantopoulos \\
  \href{http://www.let.rug.nl/alfa/}{Alfa-Informatica},
  Rijks\textit{universiteit} Groningen \\
  \href{mailto:konstant@let.rug.nl}{konstant@let.rug.nl}}

\begin{document}

\maketitle
\bibliographystyle{plainnat}

\begin{abstract}
This paper describes experiments on learning Dutch phonotactic rules
using Inductive Logic Programming, a machine learning discipline based
on inductive logical operators.
Two different ways of approaching the
problem are experimented with, and compared against each other
as well as with related work on the task. The results show a direct
correspondence between the quality and informedness of the background
knowledge and the constructed theory, demonstrating the ability of ILP
to take good advantage of the prior domain knowledge available.
Further research is outlined.
\end{abstract}

\section{Introduction}

The \term{Phonotactics} of a given language is the set of rules that
identifies what sequences of phonemes constitute a possible word in
that language. The problem can be broken down to
the \term{syllable structure} (i.e. what sequences of phonemes
constitute a possible syllable) and the processes that take place at
the syllable boundaries (e.g. assimilation).

Previous work on the syllable structure of Dutch includes hand-crafted
models~\citep{hulst:1984,booij:1995},
but also the application of machine-learning approaches such as
abduction \citep{tjong:2000} and neural networks
\citep[ch.~4]{stoianov:1998, stoianov:2001}.

This paper describes experiments on the task of constructing from
examples a Horn Clause model of Dutch monosyllabic words. The reason
for restricting the domain is to avoid the added complexity of
handling syllable boundary phonological processes. Furthermore by not
using polysyllables no prior commitment is made to any one particular
syllabification theory.

The machine learning algorithm used is described in Section~\ref{ilp},
and then the paper proceeds to describe how the training examples are
extracted from the corpus (Section~\ref{data}), what prior
knowledge is employed (Section~\ref{setup}) and, finally, draw
conclusions and outline further research (Section~\ref{concl}).

\section{Inductive Logic Programming and Aleph}
\label{ilp}

Inductive Logic Programming (ILP) is a machine learning discipline.
Its objective is the development of algorithms that construct
Predicate Logic hypotheses that can explain a set of
empirical data or observations.

The core operator used in ILP is \term{induction}.
Induction can be seen as the inverse of deduction. For example from
the clauses `All humans die' and `Socrates is human' the fact that
`Socrates will die' can be deduced. Inversely, induction uses
background knowledge (e.g. `Socrates is human') and a set of
observations (training data) (e.g. `Socrates died') to search for a
hypothesis that, in conjunction with the background knowledge, can
deduce the data. In more formal terms, given a logic program $B$
modelling the background knowledge and a set of ground terms $D$
representing the training data, ILP constructs a logic program $H$,
such that $B\land H\vDash D$ .

If the deductive operator used is resolution (as defined by
\citet{robinson:1965}), then the inductive
operator necessary to solve the equation above is the inverse
resolution operator, as defined by \citet{muggleton:1994}.

Aleph~\citep{srinivasan:aleph} is an ILP system implementing the
Progol algorithm \citep{muggleton:1995}.
This algorithm allows for single-predicate learning only, without
background theory revision or predicate invention. It incrementally
constructs the clauses of a single-predicate hypothesis that describes
the data, by iterating through the following basic algorithm:

\begin{itemise}
  
\item Saturation: pick a positive example from the training data and
  construct the most specific, non-ground clause that entails it. This
  is done by repeated application of the inverse resolution operator
  on the example, until a clause is derived that covers the original
  ground positive example and none other. This maximally specific
  clause is called the \term{bottom clause.}
  
\item Reduction: search between the maximally general, empty-bodied
  clause and the maximally specific bottom clause for a `good' clause.
  The space
  between the empty clause and the bottom clause is partially ordered
  by $\theta$-subsumption, and the search proceeds along the lattice
  defined by this partial ordering.
  The `goodness' of each clause encountered along the search path is
  evaluated by an \term{evaluation function.}
  
\item Cover removal: add the new clause to the theory and remove all
  examples covered by it from the dataset.

\item Repeat until all positive examples are covered.

\end{itemise}

The \term{evaluation function} quantifies the usefulness of each clause
constructed during the search and should be chosen so that it balances
between overfitting (i.e. covering the data
too tightly and making no generalisations that will yield coverage
over unseen data) and overgeneralising (i.e. covering the data too
loosely and accepting too many negatives).
It can be simple coverage
(number of positive minus number of negative examples covered by the
clause) or the Bayes probability that the clause is correct given the
data (for learning from positive data only) or any function of the
numbers of examples covered and the length of the clause (for
implementing bias towards shorter clauses).

\term{Syntactic bias} is applied during the reduction step to either
prune paths that are already known to not yield a winning clause, or
to enforce
restrictions on the constructed theory, for example conformance to a
theoretical framework.

\section{Training Examples}
\label{data}

As a starting point, a rough template matching all syllables is assumed.
This template is
$\mathcal{C}_3$$\mathcal{V}\mathcal{C}_5$, where $\mathcal{C}_n$
represents any consonant cluster of length up to $n$
and $\mathcal{V}$
any vowel or diphthong. The problem can now be reformulated as a
single-predicate learning task where the target theory is one of
acceptable prefixes to a given vowel and partial consonant cluster. The
rules for prevocalic and postvocalic affixing are induced in two
separate learning sessions.

The underlying assumption in formulating the problem in this fashion
is that the prevocalic and postvocalic material of a syllable is
independent from each other. In other words, it is assumed that if a
consonant cluster is acceptable as the prevocalic (postvocalic)
material of a syllable, then it is acceptable as the prevocalic
(postvocalic) material of all syllables with the same
vowel.\footnote{This assumption is consistent with hand-crafted
  syllabic models, although no claims are made here regarding its
  universality or prior necessity.}

The training data consists of 5095 monosyllabic words found in
the Dutch section of the CELEX Lexical Database, with an additional
597 reserved for evaluation. As has been mentioned above, however, the
learning task is formulated as one of learning valid affixes to a
partial syllable. In other words, the CELEX data is used to generate
the actual training examples, which are instances of a predicate
matching syllable fragments with phonemes that can be affixed at that
point.

The positive examples are constructed by breaking the phonetic
transcriptions down to three parts: a prevocalic and a postvocalic
consonant cluster (consisting of zero or more consonants) and a vowel or
diphthong. The consonant clusters are treated as \qu{affixes} to the vowel,
so that syllables are constructed by repeatedly affixing consonants, if
the \term{context} (the vowel and the pre- or post-vocalic material that has
been already affixed) allows it. So, for example, from the
word \phonemic{ma:kt} the following positives would be generated:
\begin{tabular}{ll}
\prog{prefix( m, [], [a,:] ).}     & \prog{suffix( k, [],[:,a] ).} \\
\prog{prefix( \^{}, [m], [a,:] ).} & \prog{suffix( t, [k], [:,a] ).} \\
                                   & \prog{suffix( \^{}, [tk], [:,a] ).} \\
                                   & \\
\end{tabular}
So, for example, the first two \prog{suffix} rules should be read as
follows:
\qu{\phonemic{k} can be suffixed to the \phonemic{a:} nucleus} and
\qu{\phonemic{t} can be suffixed to an \phonemic{a:k} syllable fragment}.

Note that the context lists in suffix rules are reversed, so that the
two processes are exactly symmetrical and can use the same background
predicates.

The caret, \prog{\^{}}, is used to mark the beginning and end of a word.
The reason that the affix termination needs to be explicitly licensed
is so that it is not assumed by the experiment's setup that all
partial sub-affixes of a valid affix are necessarily valid as well.

In Dutch, for
example, a monosyllable with a short vowel has to be closed, which
means
that the null suffix is not valid. The end-of-word mark allows for
this to be expressible as a theory that does not
have the following clause: \prog{suffix( \^{}, [], [V] ).}

The positives are, then, all the prefixes and suffixes
that must be
allowed in context, so that all the monosyllables in the training data
can be constructed: 11067 and 10969 instances of
1428 and 1653 unique examples, respectively.

The negative data is randomly generated words that match the
$\mathcal{C}_3$$\mathcal{V}\mathcal{C}_5$ template and do not appear
as positives.
The random generator is
designed so that the number of examples at each affix length are
balanced, in order to avoid having the large numbers of long,
uninteresting sequences overwhelm the shorter, more interesting ones.

The negative data is also split into evaluation and training data, and
the negative examples are derived from the training negative data by
the following deductive algorithm:
\begin{enumerate}
\item For each example, find the maximal substring that is provable by
  the positive \prog{prefix/3} and \prog{suffix/3} clauses in training
  data. So, for
  example, for \phonemic{mtratk} it would be \prog{trat} and
  for \phonemic{mlat}, \prog{lat\^{}}.
\item Choose the clause that should be a negative example, so that
  this word is not accepted by the target theory. Pick the inner-most
  one on each side, i.e. the one immediately applicable to the
  maximal substring computed above. For \phonemic{mlat} that would be
  \prog{suffix(m,[l],[a])}. \phonemic{mtratk}, however,
  could be negative because either
  \prog{prefix(m,[tr],[a])} or \prog{suffix(k, [t], [a])}
  are unacceptable.
  In such cases, pick one at random. This is bound to
  introduce false negatives, but no alternative that does not
  presuppose at least part of the solution could be devised.
\item Iterate, until enough negative examples have been generated to
  disprove all the words in the negative training data.
\end{enumerate}

One advantage of deriving the negative examples in this fashion
is that the importance of examples \qu{closer} to the vowel is
emphasised. This allows for a more balanced distribution of negatives
along the possible consonant cluster sizes, which would have otherwise
been flooded by the (more numerous) longer clusters.

\section{The Background Theory}
\label{setup}

Since the problem is, in effect, that of identifying the sets of
consonants that may be prefixed or suffixed to a partially constructed
monosyllable,
the clauses of the target predicate must have a means
of referring to various subsets of $\mathcal{C}$ and $\mathcal{V}$ in
a meaningful and intuitive way. This is achieved by defining a
(possibly hierarchical,) linguistically motivated partitioning of
$\mathcal{C}$ and $\mathcal{V}$. Each partition can then be referred
to as a feature-value pair, for example \feat{Lab+}
to denote the set of the labials or \feat{Voic+}
for the set of voiced consonants.
Intersections of these basic sets can then be easily referred to by
feature-value vectors; the intersection, for example, of the labials
and the voiced consonants (i.e. the voiced labials) is the
feature-value vector [\feat{Voic+},\feat{Lab+}].

The \term{background knowledge} is, as seen in section~\ref{ilp},
playing a decisive role in the quality of the constructed theory,
by implementing the theoretic framework to which the search for
a solution will be confined. In more concrete terms,
the background predicates are
the building blocks that will be used for the construction of the
hypothesis' clauses and they must
be defining all the relations necessary to discover an interesting
hypothesis.

For the purposes of this task, they have been
defined as relations
between individual phones and feature values, e.g.
\prog{labial(m,+)} or \prog{voiced(m,+)}. Feature-value vectors
can then be expressed as conjunctions like, for example,
\prog{labial(C,+)} $\land$ \prog{voiced(C,+)} to mean the voiced
labials.

Except for the linguistic features predicates, the background knowledge
also contained the \prog{head/2} and \prog{rest/2}
list access predicates. This approach was chosen over direct list
access with the \prog{nth/3} predicate,
as bias towards rules with more local context dependencies.

The theories described in sections~\ref{setup:ipa},
\ref{setup:booij} and~\ref{setup:sonority} below,
are based on background
knowledge that encodes increasingly more information about Dutch
phonology as well as Dutch phonotactics: for the experiment
in~\ref{setup:ipa} the
learner has access to the way the various symbols are arranged in the
International Phonetic Alphabet (IPA),
whereas for the experiment in~\ref{setup:booij} a
classification that is sensitive to Dutch phonological processes was
chosen.
And, finally, in section~\ref{setup:sonority}
a scalar\footnote{As opposed to binary.} sonority feature is
implemented, which has been proposed with the
explicit purpose of solving the problem of Dutch syllable structure.

The quantitative evaluation shown for three experiments was done using
the 597 words and the part of the randomly generated negative data
that have been reserved for this purpose.

\subsection{The IPA segment space}
\label{setup:ipa}
\hypertarget{setup:ipa}{}

For the first experiment
the background knowledge reflects the way that the 
International Phonetic Alphabet (IPA, 1993 version)
is organised:
the phonetic inventory of Dutch consists
of two disjoint spaces, one of consonants and one of vowels,
with three and four
orthogonal dimensions of differentiation respectively.

The consonant space varies in \term{place} and
\term{manner of articulation,} and \term{voicing.} The manner of
articulation can be \term{plosive,} \term{nasal,}
\term{lateral approximant,} \term{trill,} \term{fricative}
or \term{approximant.}
The place can be \term{bilabial,} \term{alveolar,}
\term{velar,} \term{labiodental,} \term{postalveolar}
or \term{palatal.} Voicing can be present or not.
 
Similarly for vowels, where there
are four dimensions: 
place (front, centre, back) and manner of articulation (open,
mid-open, mid-closed, closed), length and roundedness.

The end-of-word mark has no phonological features whatsoever and it
does not belong to any of the partitions of either $\mathcal{C}$ or
$\mathcal{V}$.

This schema was implemented as one background predicate per dimension
relating each phone with its value along that dimension, for example:
\begin{verbatim}
manner( plosive, p ).   place( bilabial, p ).
manner( nasal, m ).
\end{verbatim}

The evaluation function used was the
\term{Laplace function} $\frac{P+1}{P+N+2}$,
where $P$ and $N$ is the coverage of positive and negative examples,
respectively.

Since the randomly generated negatives must also contain false
negatives, it cannot be expected that even a good theory will fit it
perfectly. In order to avoid overfitting, the learning algorithm was
set to only require an accuracy of 85\% over the training data.

The resulting hypothesis consisted of 199 prefix and 147 suffix
clauses and
achieved a recall rate of 99.3\%
with 89.4\% precision.
All the false negatives were rejected because they couldn't get their
prevocalic material
licensed, typically because it only appears in a handful of loan
words. The \phonemic{\textdyoghlig} prefix necessary to accept
`jeep' and `junk', for example, was not permitted and so these two
words were rejected.

The most generic rules found were:
\begin{verbatim}
prefix(A,B,C) :- A= '^'.
prefix(A,[],C).

suffix(A,B,C) :-  A= '^'.
suffix(A,[],C).
\end{verbatim}
meaning that (a) the inner-most consonant can be anything, and (b) all
sub-prefixes (-suffixes) of a valid prefix (suffix) are also valid.

Other interesting rules include pairs like these two:
\begin{verbatim}
prefix(A,B,C) :-
   head(B,D), manner(trill,D), head(C,E), length(short,E), 
   manner(closed,E), manner(plosive,A).
prefix(A,B,C) :-
   head(B,D), manner(trill,D), head(C,E), length(short,E), 
   manner(open_mid,E), manner(plosive,A).
\end{verbatim}
and
\begin{verbatim}
prefix(A,B,C) :-
   head(B,D), manner(approx,D), head(C,E), length(short,E), 
   place(front,E), voiced(minus,A).
prefix(A,B,C) :-
   head(B,D), manner(approx,D), head(C,E), length(short,E), 
   place(front,E), manner(plosive,A), place(alveolar,A).
\end{verbatim}
that could have been collapsed if a richer feature system would
include features like `closed or mid-open vowel' and `devoiced
consonant or plosive alveolar', respectively.
These particular disjunctions might be unintuitive or even impossible
to independently motivate, but they do suggest that a redundant
feature set might allow for more interesting theories than the
minimal, orthogonal one used for this experiment. This is particularly
true for a system like Aleph, that performs no predicate invention or
background theory revision.

\subsection{Feature Classes}
\label{setup:booij}
\hypertarget{setup:booij}{}

\begin{figure}

\begin{center}
\includegraphics{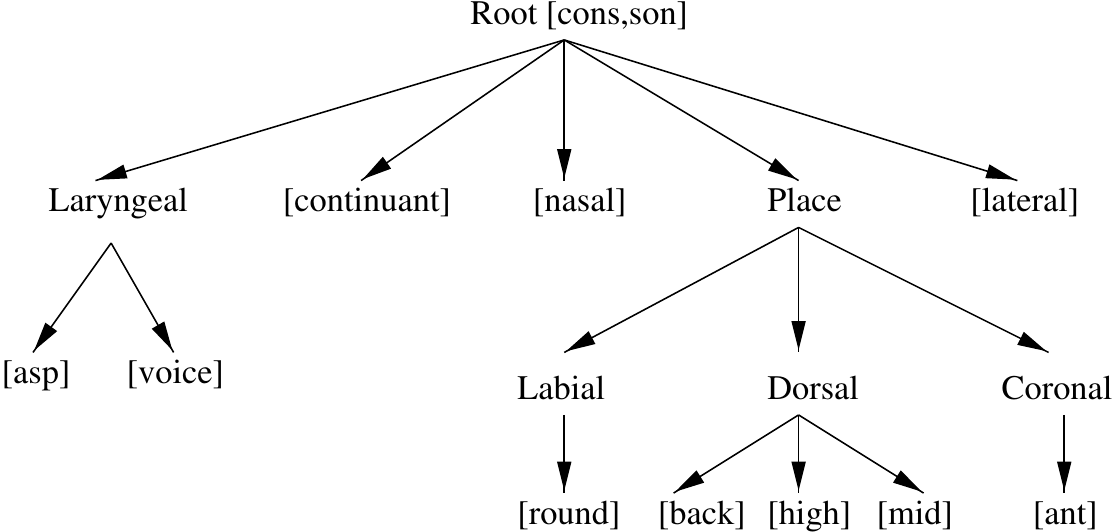}
\end{center}

\caption{The feature geometry of Dutch}
\label{fig:feat_geom}
\end{figure}

The second experiment a richer (but more
language-specific) background knowledge was made available to the
inductive algorithm, by implementing the feature hierarchy suggested
by \citet[ch.~2]{booij:1995} and shown in
figure~\ref{fig:feat_geom}.

The most generic features are the \term{major class features}
(\feat{Consonant} and \feat{Sonorant}) that are placed on the root
node and divide the segment space in
vowels [\feat{Cons-},\feat{Son+}],
obstruents [\feat{Cons+},\feat{Son-}] and
sonorant consonants [\feat{Cons+},\feat{Son+}].
Since all vowels are sonorous, [\feat{Cons-},\feat{Son-}] is an
invalid combination.

The features specifying the continuants, nasals and the
lateral \phonemic{l} are positioned directly under the root node,
with the rest of
the features bundled together under two \term{feature classes,} those
of the \term{laryngeal} and the \term{place} features. These classes
are chosen so that they collect together
features that behave as a unit in phonological processes of Dutch. The
class of laryngeal features is basically making the voiced-voiceless
distinction, while the \feat{Aspiration} feature is only separating
\phonemic{h} from the rest. The \term{place} class bundles
together three subclasses of place of articulation features, one for
each articulator.
Furthermore some derived or redundant features such as \feat{Glide,}
\feat{Approximant} and \feat{Liquid} are defined, but not shown in
Figure~\ref{fig:feat_geom}. The vowels do not
include the schwa, which is set apart and only specified as
\feat{Schwa+}.

Using the Laplace evaluation function and this background knowledge,
the
constructed theory consisted of 13 prefix and 93 suffix rules, 
accepting 94.2\% of the test positives and under 7.4\% of the test
negatives.

Among the rejected positives are loan words (`jeep' and `junk' once
again), but also all the
words starting with perfectly Dutch
\phonemic{s} - obstruent - liquid clusters.

The prefix rule with the widest coverage is:
\begin{verbatim}
prefix(A,B,C) :-
   head(C,D), sonorant(D,plu), rest(B,[]).
\end{verbatim}
or, in other words, `prefix anything before a single consonant
before a nucleus other than the schwa'. 

The suffix rules were less strict, with only 3 rejected positives,
`branche', `dumps' and `krimpst' (the first two of which are loan
words) that failed to suffix
\phonemic{\textesh}, \phonemic{s} and \phonemic{s} respectively.
Some achieve wide coverage (although never quite as wide as that of
the prefix rules,) but some are making reference to individual phonemes
and are of more restricted application. For example:
\begin{verbatim}
suffix(A,B,C) :-
   rest(C,D), head(D,E), rest(B,[]), A=t.
\end{verbatim}
or, `suffix a \phonemic{t} after exactly one consonant, if the nucleus
is a long vowel or a diphthong'.

Of some interest are also the end-of-word marking rules
(see in section~\ref{setup} above about the \prog{\^{}} mark),
because of the
fact that open, short monosyllables are very rare in Dutch (there
are four in CELEX: `schwa', `ba', `hè', and `joh'). This would suggest
that the best way to treat those is as exceptions, and have the
general rule disallow open, short monosyllables. What was learned
instead was a whole set of 29 rules for suffixing~\prog{\^{}}, the most
general of which is:
\begin{verbatim}
suffix(A,B,C) :-
   head(B,t), larynx(t,E), rest(B,F), 
   head(F,G), larynx(G,E), A= '^'.
\end{verbatim}
or `suffix an end-of-word mark after at least two consonants, if the
outer-most one is a \phonemic{t} and has the same values
for all the
features in the \feat{Laryngeal} feature class as the consonant
immediately preceding it'.

A final note that needs to be made regarding this experiment, is one
regarding its computational complexity. Overlapping and redundant
features might be offering the opportunity for more interesting
hypotheses, but are also making the search space bigger. The reason is
that overlapping features are diminishing the effectiveness of the
inverse resolution operator at keeping uninteresting predicates out of
the bottom clause: the more background predicates can be used to prove
the positive example on which the bottom clause is seeded, the longer
the latter will get.

\subsection{Sonority Scale}
\label{setup:sonority}
\hypertarget{setup:sonority}{}

This last experiment implements and tests the syllabic structure model
suggested by \citet[ch.~3]{hulst:1984}. This is not a machine learning
experiment, but a direct implementation of a hand-crafted syllable
structure theory, and the presented for the sake of comparison.
The Dutch syllable is
analysed as having three prevocalic and 5 postvocalic positions (some
of which may be empty), and constraints are placed on the set of
consonants that can occupy each.

\begin{table}
\begin{center}
\begin{tabular}{|l|ccccccc|}
\hline
phoneme  & obstruents & m & n    & l   & r    & glides & vowels \\
sonority &   1        & 2 & 2.25 & 2.5 & 2.75 & 3      & 4      \\
\hline
\end{tabular}
\caption{The Sonority Scale}
\label{sonority:scale}
\end{center}
\end{table}

The most prominent constraint is the one stipulating a high-to-low
\term{sonority} progression from the nucleus outwards.
The theory is that
each phoneme is assigned a sonority value
(as in table~\ref{sonority:scale})
and syllables are then built from the nucleus
outwards, by stacking segments of decreasing sonority.

The rough outline of the sonority scale is based on
language-independent characteristics of the segments themselves and
will, for example, place vowels higher than consonants and obstruents
lower that any other consonant. The scale shown here for Dutch
has been further refined in
such a way that particular idiosyncrasies of the language are
predicted.

So, for example, the nasals and the liquids are given in the original
(language-independent) approximation of the sonority scale a sonority
of 2. In the final, complete theory, however, they are evenly spread
along the space between 2 and 3. The justification for this
refinements is to allow for distinctions
among the nasals and the liquids, so that, for example,
\phonemic{karl} would be acceptable but \phonemic{kalr} not.
It must, therefore, be noted that the
prior knowledge of this theory is not only
language-specific, but is also directly aimed at solving the very
problem that is being investigated. In other words, it is a very
informed piece of prior knowledge that, at least partially,
encodes a hand-crafted model of Dutch syllable structure.

In addition to the high-to-low sonority level progression from the
nucleus
outwards, there are both \term{filters} and explicit licensing rules.
The former are restrictions referring to sonority
(e.g. `the sonority of the three left-most positions must be smaller
than 4')
or other phonological features
(e.g. the `no voiced obstruents after the vowel' filter in p.~92)
and are applicable in conjunction to the sonority rule.
The latter are typically restricted in scope rules, that take
precedence over the sonority-related constraints mentioned so far.
The left-most position, for
example, may be \phonemic{s} or empty, regardless of the contents of
the rest of the prevocalic material.

Implementing the basic sonority progression rule as well as the most
widely-applicable filters and rules\footnote{Some were left out
  because they were too lengthy when translated from their
  fixed-position framework to the affix licensing one used here, and
  were also very specifically fine tuning the theory to individual
  consonant clusters.}
yielded impressive compression rates
matched with results lying between those of the two previous
experiments: 93.1\% recall, 83.2\% precision.

\section{Conclusions and Further Work}
\label{concl}

The quantitative results from the machine learning experiments
presented above are collected in table~\ref{conclusions:results},
together with those of \citet{tjong:2000}\footnote{From experiments on
  phonetic data in the `Experiments without Linguistic Constraints'
  section.}
and the results from the sonority scale experiment.
Those last ones in particular, are listed
for comparison's
sake and as the logical end-point of the progression towards more
language- and task-specific prior assumptions.

The first two columns are directly comparable, because they are both
only referring to phonetic primitives with no linguistically
sophisticated background knowledge. The fact that the
$\mathcal{C}_3$$\mathcal{V}\mathcal{C}_5$ template assumed in this
work is not taken for granted in \citep{tjong:2000}, is compensated in
terms of 
compactness as well as performance.
Compactness because the numbers of rules quoted in the first column do
not include the 41 extra rules (besides the 1154 prefix and
suffix rules) that describe the "basic word" on which the affix rules
operate. Performance because 
the precision given is measured on totally
random strings, whereas in this work only strings matching the
$\mathcal{C}_3$$\mathcal{V}\mathcal{C}_5$ template are used.

As can be seen, then, the ILP-constructed rules compare favourably (in
both performance and hypothesis compactness) with those constructed by
the deductive approach employed in \citep{tjong:2000}.

\begin{table}
\begin{center}
\begin{tabular}{|r|l|l|l|l|}
\hline
             & (Tjong 2000) & \hyperlink{setup:ipa}{IPA}
                                        & \hyperlink{setup:booij}{Feat. Classes} 
                                                    & \hyperlink{setup:sonority}{Sonority} \\
\hline
Recall       & 99.1\%       & 99.3\%    & 94.2\%    & 93.1\%   \\
Precision    & 74.8\%       & 79.8\%    & 92.6\%    & 83.2\%   \\
Num. Clauses & $577+577$    & $145+36$  & $13+93$   & 3+8      \\
\hline
\end{tabular}
\caption{Results}
\label{conclusions:results}
\end{center}
\end{table}

What can be also seen by comparing the two ILP results
with each other, is that the drop in recall between the
the second and third column is compensated by higher precision and
compression, suggesting
a direct correspondence between the quality of the
prior knowledge encoded in the background theory and that of the
constructed hypothesis.

The strongest argument in favour of ILP is that is employing
background knowledge expressed in a symbolic formalism, allowing for
easy transfer of prior domain knowledge into the process. This is
consistent with the results shown presented here, since they
demonstrate that the ILP algorithm is taking advantage of
the more informed and complex background theory of
feature classes to construct a better and shorter theory.


One interesting follow-up to these experiments would be attempting to
expand their domain to that of syllables of polysyllabic words and,
eventually, full word-forms. In the interest of keeping the problems
of learning syllabic structure and syllable-boundary phonology apart, a way
must be devised to derive from the positive data (i.e. a corpus of
Dutch word-forms) examples for a distinct machine learning session for
each task.

Furthermore, it would be interesting to carry out the same (or rather
the equivalent) experiments on other parts of the CELEX corpus (e.g.
English or German) and see to which extend the results-to-background
relation follows the same patterns.

\bibliography{cs,ling,logic,url}

\end{document}